# A metaheuristic multi-objective interaction-aware feature selection method

Motahare Namakin[1], Modjtaba Rouhani[1] and Mostafa Sabzekar[2,*]

**Abstract-** Multi-objective feature selection is one of the most significant issues in the field of pattern recognition. It is challenging because it maximizes the classification performance and, at the same time, minimizes the number of selected features, and the mentioned two objectives are usually conflicting. To achieve a better Pareto optimal solution, metaheuristic optimization methods are widely used in many studies. However, the main drawback is the exploration of a large search space. Another problem with multi-objective feature selection approaches is the interaction between features. Selecting correlated features has negative effect on classification performance. To tackle these problems, we present a novel multi-objective feature selection method that has several advantages. Firstly, it considers the interaction between features using an advanced probability scheme. Secondly, it is based on the Pareto Archived Evolution Strategy (PAES) method that has several advantages such as simplicity and its speed in exploring the solution space. However, we improve the structure of PAES in such a way that generates the offsprings, intelligently. Thus, the proposed method utilizes the introduced probability scheme to produce more promising offsprings. Finally, it is equipped with a novel strategy that guides it to find the optimum number of features through the process of evolution. The experimental results show a significant improvement in finding the optimal Pareto front compared to state-of-the-art methods on different real-world datasets.

## 1. Introduction

Feature selection (FS) is one of the most important problems and an active research area in pattern recognition, such as classification. Many datasets in the real-world include irrelevant, redundant, or even noisy features, which adversely affect the classification efficiency. Thus, the main aim of a FS approach is to find the optimal subset of features by eliminating such less informative ones. Some major benefits of FS are including reducing the computational complexity of training classifiers, avoiding over-fitting, and as a result, improving the interpretability of the final trained model, and increasing the classification performance by eliminating the irrelevant and redundant features. But, with no prior knowledge about a given dataset, it is difficult to find informative features. Furthermore, the other reasons that caused feature selection to be a challenging problem can be summarized as follows. Firstly, FS is an NP-hard problem due to the large search space. With increasing the number of features, the

Mostafa Sabzekar
sabzekar@birjandut.ac.ir

[1] Department of Computer Engineering, Ferdowsi University of Mashhad, Mashhad, Iran
[2] Department of Computer Engineering, Birjand University of Technology, Birjand, Iran

search space size is grown exponentially. The second reason is related to the correlation between features that can directly affect classification accuracy. In this case, an individually informative feature may become irrelevant when used with other features and vice versa.

Up to now, many efforts have been made to deal with the FS challenges. These studies can be categorized based on different aspects such as evaluation criteria, searching technique, and the number of objectives. Based on the evaluation criteria, the presented approaches in the literature can be broadly categorized into the filter, wrapper, and embedded methods. The first category, namely filtering techniques, achieves the best feature subset based on a predefined evaluation criterion, which is based on the characteristics of the features and is considered independent of the learning model. Although these methods are faster and more scalable, they suffer from lower accuracy than the approaches in other categories. The wrapper methods utilize the classification performance as evaluation criteria for each candidate feature subset. As a result, they have high time complexity but provide more accurate results than the filter methods. Finally, the embedded approaches find the optimal solution during the training process. They benefit from the advantages of both filter and wrapper methods. However, the main drawback of these methods is that they limit to specific learning models.

Based on searching techniques of the solution space, the FS methods can be generally divided into the exhaustive search, heuristic search, and random search [2]. The first group techniques search the entire problem space. These methods guarantee to find the global optimum, but they cannot be applied to solve a vast range of FS problems. On contrary, there are many researches utilize the heuristic search techniques such as greedy search methods. The two most popular methods in this group are sequential backward elimination and sequential forward selection. However, the main weaknesses of these techniques are that they firstly may easily trap in a local optimum. Secondly, they suffer from the "nesting effect", namely, once a feature is eliminated or selected, we cannot select or eliminate it in the next steps. Lastly, the random search methods make a trade-off between the two later mentioned categories. It means that they have the capability of global searching and also reasonable time complexity. Over the past decade, metaheuristic methods have provided promising results when they come to the feature selection problems. Among them, genetic algorithm (GA), ant colony optimization (ACO), particle swarm optimization (PSO), differential evolution (DE), artificial bee colony (ABC), and firefly algorithm have received more attention from the scientific community [19]. The main strengths of these methods are that they need no assumption or any domain knowledge about the problem, and also, these algorithms can create multiple solutions in each run. However, the main challenge about them is exploration of the large search space of the problem that causes the high time complexity problem.

Regarding the number of objectives, the FS methods are categorized into single-objective (SO) and multi-objective (MO) methods. The SO methods consider only one objective, namely classification accuracy, as the fitness function. However, the MO methods try to minimize the number of selected features in addition to maximizing the classification accuracy. The main issue about the MO approaches is that these two goals are usually conflicting. In the next section, the recent studies about the multi-objective FS problem are surveyed in more detail.

To tackle the limitations of the FS methods, we propose a multi-objective algorithm that combines the Pareto Archived Evolution Strategy (PAES) [12] with our recently introduced estimation of distribution algorithm (EDA)-based approach (SC-FS) that can consider the interaction between features [18]. PAES is a multi-objective evolutionary algorithm that uses single parent and single offspring for the evolution process. Thus, it can be an appropriate choice for solving feature selection problems due to its simplicity and speed in exploring the solution space. However, to create an offspring from a parent, (1+1)-PAES uses only a blind mutation as the evolutionary operator, and consequently, the offspring may have a lower fitness value than the parent. The proposed method is designated based on the PAES powered by our SC-FS to obtain better Pareto front solutions. Each offspring is generated using our proposed update procedure. Thus, it will be capable to consider the interaction between features. Indeed, instead of using a blind mutation, we utilize a probability scheme to generate more promising individuals. Furthermore, we introduce a guiding strategy to determine the number of selected features for each individual. Consequently, the proposed algorithm selects the optimum number of features. Thus, we can summarize the main contributions of this paper as follows:

- *Dealing with the complementary features*: it can deal with the significance of each feature and also their interactions with two introduced data structures *SV* and *IM*. They are updated after finding the *winner* and the *loser* in each iteration of the proposed algorithm.
- *Guiding strategy*: it is equipped with a novel strategy that guides it to find the optimum number of features. The number of features for each offspring is calculated by the $\chi^2$-distribution.
- *Intelligent generating of offsprings:* it considers both the importance degrees of each feature and correlated ones, which are lie in the introduced conditional probabilities. It should be noted that each feature of offsprings is selected with a roulette wheel using these probabilities. Therefore, we enriched the PAES with an intelligent generating of offsprings in comparison to a blind mutation.

The rest of the paper is organized as follows. Section 2 provides a brief review of the recently published metaheuristic multi-objective feature selection methods. The structure of the proposed algorithm is described in section 3. Then, in section 4, the performance of our proposed method is investigated. Finally, section 5 brings some concluding remarks followed by suggestions for future studies.

## 2. Literature review

The metaheuristic optimization methods can be generally studied into single-objective and multi-objective methods. In many researches, the FS has been modeled as a single-objective optimization problem that maximizes only the classification accuracy, but it has a multi-objective nature. To find the best subset of features, we should maximize the classification performance and simultaneously minimize the number of selected features. This is a challenging problem because the two mentioned objectives are usually conflicting [31]. Hence, in multi-objective optimization problems (MOOP), a set of non-dominated solutions, known as the Pareto front, is returned as the solution.

There are many efforts in the literature that treats the FS problem as a MOOP. Nowadays, metaheuristic algorithms have been applied widely for finding the best feature subset. They can explore the search space to obtain multiple solutions on the Pareto optimal front in a single run [20]. Multi-objective feature selection also has been widely used in different applications such as online sales forecasting [10], software defect prediction [21], drug's dose prediction [24], intrusion detection [32], time series prediction [1], credit scoring [13], [23], medicine [17], etc. In the following, we study only the recently published metaheuristic-based feature selection studies. It is worthy of mentioning that there are several ways to classify these researches. Here, these studies are categorized and studied based on their evaluation criteria.

### 2.1. Filter approaches

In [3], a PSO-based multi-objective feature selection approach (called RFPSOFS) has been developed. Its main contribution is the simultaneous use of objective and problem spaces. To do this, firstly, the features were ranked based on their frequency in the archive, and then the ranked features were used to improve the quality of solutions in the archive set by moving the particles more purposefully. The elitism level of the algorithm was increased, but it can be controlled by the mutation operator.

Wang et al. [27] introduced a MO approach (called MECY-FS) based on new feature redundancy and correlation metrics. They tried to find a compact solution with minimum redundancy and maximum correlation.

### 2.2. Wrapper approaches

A multi-objective feature selection approach that utilized the binary differential evolution (DE) and self-learning (called MOFSBDE) was proposed in [31]. Three new operators were introduced to improve its performance. The first operator steers individuals rapidly towards potentially better solutions. The second operator enhances the self-learning capabilities of the elite individual. The last operator stands for the selection operator of DE to reduce its computational complexity.

Niu et al. [22] presented a bacterial foraging optimization (BFO) based multi-objective feature selection method, called MOBIFS. The redundant features were removed using a wheel roulette mechanism. To avoid trapping into the local optima, four information exchange strategies were integrated into the BFO algorithm. The proposed strategy helps the individuals to exchange their helpful information. Furthermore, the KNN classification algorithm was used for the evaluation of each subset of features.

An improved version of the cat swarm optimization (CSO) method was developed in [8] (called HCSO). The conventional CSO algorithm was enhanced by some introduced characteristics, namely guided, competitive, and inherent. The overall performance of the CSO was improved by these techniques. However, the time complexity of the presented method remained unchanged.

To model the FS as a multi-objective problem, the authors in [9] were benefited from the artificial bee colony (ABC) algorithm. They combined ABC with non-dominated sorting and also genetic operators. They implemented both binary and continuous representations of the ABC and investigated their proposed algorithms. In another research, Zhang et al. [29] proposed a cost-sensitive FS algorithm using a multi-objective ABC method, called TMABC-FS. They introduced two new operators and two archives. The two new operators were convergence and diversity guiding search strategies which were applied to the employed and the onlooker bees, respectively. The first operator used the social learning capability of the particle swarm optimization (PSO) algorithm to increase the convergence speed of the proposed method. The second operator improves the distribution of Pareto optimal solutions. Additionally, to ensure the diversity of non-dominated solutions both in objective and variable spaces. Similarly, the ABC algorithm is used in [26] to solve multi-objective FS problems. The proposed method introduced two strategies, namely a sample utilization and K-means-based differential selection to reduce the sample size of ABC. Besides, it improved the classification efficiency by providing an elite search operator for employed bees.

Teaching Learning Based Optimization (TLBO) is another metaheuristic method used to solve the problem in [11]. Three different versions of multi-objective TLBO were proposed such that there is no need to tune any parameter during the optimization.

In another study [20], a PSO-based MO technique, called ISRPSO, was presented to achieve a set of non-dominated solutions with higher classification efficiency using local search techniques. Zhang et al. [30] enhanced the search capability of multi-objective PSO by combining multiple ideas, namely a hybrid operator, the crowding distance, two archive sets, and Pareto domination relations. The proposed method (called HMPSOFS) can explore the search space more effectively to achieve a set of nondominated feature subsets.

## 2.3. Hybrid approaches

In [14], a hybrid filter-wrapper method was developed for gene selection using simplified swarm optimization, called MOSSO. An emerging aggregate filter approach was combined with a wrapper strategy to obtain the best feature subset. Furthermore, a weighting scheme is introduced for guiding the method towards the interesting regions.

The authors in [16] combined the filter and wrapper strategies for multi-objective feature selection. In their method, called GRMOEA, the NSGA-II was adopted as the framework of both wrapper and filter populations. Then, the two populations were evolved simultaneously using *guiding* and *repairing* strategies. The first one used the good solution in wrapper population to guide the filter population towards a better direction. The latter one repaired some features of wrapper population by the good features determined by the filter population. Table 1 summarizes the mentioned studies for better comparison.

Table1. Comparison of the selected metaheuristic multi-objective FS methods.

| Studies | Type | Metaheuristic method | Evolution measure | Objectives |
|---|---|---|---|---|
| RFPSOFS [3] | Filter | PSO | Feature ranking | Minimization of: 1) attribute number, and 2) classification error |
| MECY-FS [27] | Filter | MECY | Mutual Information | Minimization of: 1) Cluster quality, and 2) classification error |
| MOFSBDE [31] | Wrapper | DE | KNN | Minimization of: 1) attribute number, and 2) classification error |
| MOBIFS [22] | Wrapper | BFO | KNN | Minimization of: 1) attribute number, and 2) classification error |
| HCSO [8] | Wrapper | CSO | SVM | Minimization of: 1) attribute number, and 2) classification error |
| Hancer et al. [9] | Wrapper | ABC | KNN | Minimization of: 1) attribute number, and 2) classification error |
| TMABC-FS [29] | Wrapper | ABC | KNN | Minimization of: 1) feature cost, and 2) classification error |

| FMABC-FS [26] | Wrapper | ABC | KNN | Minimization of: 1) attribute number, and 2) classification error |
|---|---|---|---|---|
| Kiziloz et al. [11] | Wrapper | TLBO | LR, ELM, SVM | Minimization of: 1) attribute number, and 2) classification error |
| ISRPSO [20] | Wrapper | PSO | KNN | Minimization of: 1) attribute number, and 2) classification error |
| HMPSOFS [30] | Wrapper | PSO | KNN | Minimization of: 1) feature cost, and 2) classification error |
| MOSSO [14] | Hybrid | SSO | Emerging aggregate filter, SVM | Minimization of: 1) attribute number, and 2) classification error |
| GRMOEA [16] | Hybrid | NSGA-II | KNN | • Wrapper: Minimization of: 1) attribute number, and 2) classification error<br>• Filter: Minimizing the redundancy between features and maximizing the relevance between them. |

By reviewing the literature, we can conclude that the metaheuristic multi-objective feature selection techniques have two main goals: first, decreasing the time complexity, and second, enhancing the search capability to obtain better non-dominated feature subsets. Various operators were introduced to achieve these goals. However, it is still an open issue due to its wide usage in different applications and also the demand for methods with more and more accurate Pareto optimal solutions and lower complexity of time. Another big challenge in any FS problem is the features interactions. A given feature may be relevant (weakly relevant) to the target class individually, but the classification accuracy is decreased (improved) when it used with other features. In the other words, two individually relevant features may be highly dependent on each other, so just one of them might be enough to improve the accuracy. One of the main advantages of our proposed multi-objective FS method can meet this challenge. In the next section, we propose our method that not only considers the interaction between features, but also is benefited by the PAES that uses only one parent and one offspring in each iteration and therefore is quite fast.

## 3. The proposed method

To tackle the mentioned challenges of multi-objective FS methods, we present a metaheuristic multi-objective interaction-aware feature selection (MMI-FS) approach. It is based on the Pareto Archived Evolution Strategy (PAES) enriched by our proposed SC-FS. Recently, we proposed a single-objective correlation-aware FS (SC-FS) method in [18] that deals with both importance of each feature and the interaction between them. In the present study, our proposed SC-FS is integrated with the PAES optimization algorithm to find more promising Pareto front optimal solutions. For a better illustration of the proposed method framework, firstly, let us briefly review the structure of PAES.

## 3.1. Pareto archived evolution strategy (PAES)

The Pareto archived evolution strategy (PAES) was introduced by Knowles and Corne [12]. It suggests a multi-objective evolutionary algorithm (MOEA) based on (1+1)-evolution strategy that reduces its computational cost [4]. A single parent and single offspring are also used for the evolution process. Binary representation and bitwise mutations are employed for creating the offsprings. In each iteration of PAES, the offspring $O$ is compared with the parent $P$. If $P$ is dominated by $O$, the offspring $O$ is approved and it is considered as the new parent. Then, the algorithm goes to the next iteration. In contrast, if $P$ dominates $O$, the algorithm rejects $O$, and a new offspring is created using the mutation operator. Finally, if none of $P$ and the $O$ cannot dominate each other, choosing between $P$ and the $O$ is made by the best solution in the archive. In this situation, $O$ is compared with solutions in the archive. If $O$ dominates some of them, it is approved as the new parent, and all of the dominated members are removed from the archive set. If $O$ cannot dominate any member of the archive, $P$ and $O$ are investigated for their closeness to members of the archive. Then, $O$ is approved and considered as the new parent if it is placed in a less crowded region. The PAES is used to solve optimization problems in various applications because it presents a very straightforward and fast baseline approach for multi-objective optimization problems. The main advantages of (1+1)-PAES can be summarized as its high speed, coverage of the optimal Pareto front, and the general absence of strong bias [12]. Despite these advantaged, (1+1)-PAES uses only the mutation operator is used for creating new offsprings in (1+1)-PAES. Therefore, the offspring may be inferior to the parent.

## 3.2. The proposed metaheuristic multi-objective interaction-aware feature selection method (MMI-FS)

In this paper, we propose a metaheuristic multi-objective feature selection method, based on the PAES enriched by our proposed SC-FS. The proposed SC-FS is based on the estimation of distribution algorithm (EDA). The EDA algorithm generates two individuals in each iteration. Then, they compete and determine the *winner* and *loser* individuals. To deal with the correlated features, we defined two data structures, namely *significance vector (SV)* and *interaction matrix (IM)*, to implement the effects of both each feature alone and interaction between them. As a result, the classification performance of our proposed algorithm will be improved. Let $n$ be the number of original features of a given dataset. The $SV(i)$ with size $n$ denotes the goodness of the $i$-th feature and $IM(i,j)$ represents the goodness of appearing both features $i$ and $j$ in the optimal subset.

To generate two independent individuals, the first and the $d$-th feature for each them is chosen by the roulette wheel selection strategy using the following equations:

$$P(X_j^1) = \frac{SV(j)}{\sum_{k=1}^{n} SV(k)}, \forall j = 1,...,n, \qquad (1)$$

$$P(X_j^d \mid X_l \in A) = \frac{\left(\prod_{X_l \in A} IM(X_j, X_l)\right) \times SV(X_j)}{\left(\sum_{X_z \in A} \prod_{X_l \in A} IM(X_z, X_l) \times SV(X_z)\right)}, \quad \forall k = 2,...,s, \qquad (2)$$

Based on the conditional probabilities in Equations (1), a greater value gives higher chance to the *j*-th feature of individual *X* ($X_j$) to be chosen as the first feature. Besides, Equation (2) calculates the probability of the *j*-th feature to be chosen as the *d*-th feature of *X*, when a subset of feature *A* is chosen in advance. It should be mentioned that $\bar{A}$ denotes the features that have not been selected yet by the algorithm. Then, the generated individuals compete based on the fitness function to determine the *winner* and *loser* individuals. The *SV* and *IM* data structures are then updated based on the *winner* and *loser* using our update procedure described in Table 1.

Table 1. The update procedure of *SV* and *IM*

| $l_i, l_j$ \ $w_i, w_j$ | 0,0 | 0,1 | 1,0 | 1,1 |
|---|---|---|---|---|
| **0,0** | $IM(i,j)$, $SV(i)$, $SV(j)$ | $IM(i,j)$, $SV(i)$, $SV(j)-\alpha$ | $IM(i,j)$, $SV(i)-\alpha$, $SV(j)$ | $IM(i,j)-\alpha$, $SV(i)-\alpha$, $SV(j)-\alpha$ |
| **0,1** | $IM(i,j)$, $SV(i)$, $SV(j)+\alpha$ | $IM(i,j)$, $SV(i)$, $SV(j)$ | $IM(i,j)$, $SV(i)-\alpha$, $SV(j)+\alpha$ | $IM(i,j)-\beta$, $SV(i)-\alpha$, $SV(j)$ |
| **1,0** | $IM(i,j)$, $SV(i)+\alpha$, $SV(j)$ | $IM(i,j)$, $SV(i)+\alpha$, $SV(j)-\alpha$ | $IM(i,j)$, $SV(i)$, $SV(j)$ | $IM(i,j)-\beta$, $SV(i)$, $SV(j)-\alpha$ |
| **1,1** | $IM(i,j)+\alpha$, $SV(i)+\alpha$, $SV(j)+\alpha$ | $IM(i,j)+\beta$, $SV(i)+\alpha$, $SV(j)$ | $IM(i,j)+\beta$, $SV(i)$, $SV(j)+\alpha$ | $IM(i,j)$, $SV(i)$, $SV(j)$ |

As shown in Table 1, some updates on *SV* and *IM* will be performed based on the different values of features *i* and *j* for the winner and the loser. The parameter *α* is a positive value between zero and one. It stands for weakening or strengthening the significance value of the *i*-th, namely *SV(i)*, and also, the goodness of selecting both features *i* and *j*, namely *IM(i,j)*. The parameter *β* has a greater value (i.e., two or three times higher) than *α*, and used in the states that we have more confidence in simultaneous selection of features *i* and *j*.
When we are more confident in decreasing or increasing the chance of selecting the features.

Another advantage of our proposed MMI-FS is that it utilizes a *guiding strategy*. In this strategy, for each individual, the number of selected features is calculated using a $\chi^2$-distribution with *r* degree of freedom, where *r* is the number of selected features by the *winner* in each iteration of the algorithm. Consequently, the best value for the number of features is determined by the evolution process.

As mentioned before, to deal with multi-objective problems, the proposed algorithm is based on PAES. PAES is one of the most promising solutions for solving multi-objective optimization problems. It suggests (1+1)-ES that reduces the time complexity of finding the optimal Pareto solutions. However, since the PAES generates only one offspring in each iteration, using only a blind mutation as the reproduction operator can easily affect the quality of the explored solutions. To overcome this problem, we utilize the above introduced probability scheme to generate more promising offspring in each iteration. Therefore, the proposed MMI-FS is equipped with an intelligent generating process of new offspring compared with a blind one in PAES. The pseudocode of the proposed algorithm is described in Algorithm 1.

**Algorithm 1: The structure of MMI-FS**

```
1    Generate random parent P;
2    Evaluate (P);
3    AddToArchive(P);
4    while (not (terminationCondition ())) do
5        Generate new offspring O;   //Algorithm 2
6        if (P dominate O)
                Reject O;
                winner ← O; loser ← P;
                Update PV and PM;   //Table 1
         else
             Compare O with all members of the archive set;
7            if (O dominated by any member of the archive)
8                Reject O;
9                winner ← P; loser ← O;
10               Update PV and PM;   // Table 1
11           else
12               if (O dominates any member of the archive)
13                   Eliminate all dominated members;
14                   AddToArchive(O);
15                   Approve O; P ← O;
16                   winner ← O; loser ← P;
17                   Update PV and PM;   // Table 1
18               else
19                   if (isFull(archive))
20                       diversityCheck(O);
21                   else
22                       AddToArchive(O);
23                   end if
24                   if (O places in a less crowded region than P)
25                       Approve O;  P ← O;
26                       winner ← O; loser ← P;
27                       Update PV and PM;   // Table 1
28                   else
29                       Reject O;
30                       winner ← O; loser ← P;
31                       Update PV and PM;   //Table 1
32                   end if
33               end if
34           end if
39       end if
40   end while
```

The algorithm starts with generating a random parent and adds it to the archive set. Unlike the PAES, which performs a simple mutation to generate a new offspring, the proposed method considers our proposed conditional probabilities for this purpose as described in Algorithm 2. This improves the searching capability of the proposed approach to find a better Pareto front solution. The parent $P$ and the offspring $O$ are then compared based on two objectives, namely maximizing the classification performance, and at the same time, minimizing the number of selected features. Based on the obtained result from the comparison, different scenarios occur as follows.

1) If *P* dominates *O*, the algorithm rejects *O* and generates a new offspring. *P* and *O* are considered as the *winner* and the *loser*, respectively. Then, the *winner* and the *loser* are used for updating the structures *SV* and *IM* as shown in Table 1.
2) If *P* does not dominate *O*, the offspring is compared with other members of the archive set. In this case, the following three situations may occur:
   a. If any member of the archive set dominates *O*, the algorithm rejects it, defines the *winner* (*P*) and *loser* (*O*), and updates *SV* and *IM*.
   b. If *O* dominates any member of the archive set, the algorithm eliminates all dominated members from the archive set, adds *O* to the archive set, considers *P* and *O* as the *loser* and the *winner*, respectively, and updates *SV* and *IM*. It should be noted that in this situation, the parent *P* has not been able to dominate any member of the archive set. However, the offspring *O* dominates some members of the archive set. For this reason, we can conclude the offspring *O* and the parent *P* are considered as the *winner* and *loser*, respectively.
   c. If *O* and the members of the archive set cannot dominate each other, the algorithm checks two conditions: 1) whether the archive set is full; and 2) whether *P* or *O* is located in a less crowded region. Firstly, if the archive set is not full, *O* is added to the archive set. However, if the archive set is full, the algorithm checks whether *O* increases the degree of diversity or not. If it increases the diversity, it will be replaced with the members in the most crowded location. Secondly, if *O* resides in a less crowded region than *P*, the algorithm accepts the offspring *O* and otherwise, rejects it. The *winner* and the *loser* are then identified and the *SV* and *IM* are updated using them.

Algorithm 2 describes the steps of generating a new offspring at the begging of each iteration (line 5, Algorithm 1).

| | **Algorithm 2: Generating a new offspring** |
|---|---|
| 1 | Select the first feature of offspring *O* with roulette wheel using probabilities *P (Equation (1))*: $$P(O_j^1) = SV(j) / \sum_{k=1}^{n} SV(k);$$ |
| 2 | Determine the number of selected features for *O* using $\chi^2$-distribution with *r* degrees of freedom: $$m \sim \chi^2(r);$$ |
| 3 | **for** *d* =2 to *m* |
| 4 | Determine *d*-th feature for *O* with roulette wheel using probabilities *P (Equation (2))*: $$P(O_j^d \mid O_l \in A) = \frac{\left(\prod_{X_l \in A} IM(O_j, O_l)\right) \times SV(O_j)}{\left(\sum_{X_z \in A} \prod_{X_l \in A} IM(O_z, O_l) \times SV(O_z)\right)};$$ |
| 5 | **End for** |

Thus, we can summarize the main specifications of our proposed multi-objective FS method (MMI-FS) as follows: firstly, the proposed method can consider the importance of each feature alone, and simultaneously, the feature interaction using two data structures *SV* and *IM*. They are updated after identifying the *winner* and the *loser* in each iteration of the algorithm (see Table 1). Additionally, they determine the probability of selecting each feature for the offsprings (see Equations (1) and (2)). Secondly, it is equipped with a novel strategy to find the optimum number of features. The number of features for each offspring (the parameter *m* in line 2, Algorithm 2) is defined by $\chi^2$-distribution with *r* degrees of freedom. The parameter *r* is the feature number of the *winner*, which will be updated and optimized through the evolution of the algorithm. Finally, it generates the offsprings according to Algorithm 2. Thus, we consider both the importance degrees of each feature and correlated ones, which lie in the introduced conditional probabilities. It should be noted that each feature of the offsprings is selected with a roulette wheel using these probabilities. Therefore, we enrich the PAES with an intelligent generating of offsprings. In the next section, we evaluate our proposed MMI-FS approach and compare it with the state-of-the-art methods on different standard benchmark datasets.

## 4. Experimental results

In this section, we compare the performance of our multi-objective feature selection method (MMI-FS) with both basic and state-of-the-art MOOP methods using various datasets. The methods used for comparisons are NSGA-II [5], MOFSBDE [31], and GRMOEA [16]. NSGA-II is one of the most efficient and popular multi-objective metaheuristic algorithms that adopts crowding distance and non-dominated sorting strategies. It benefits from the high convergence speed to achieve a more accurate solution set [15]. The other two methods, namely MOFSBDE and GRMOEA, were described in section 2. For a fair comparison, the datasets in all experiments are separated randomly into 75% training, and 25% testing data. The details of 12 datasets obtained from the UCI Repository [6] are summarized in Table 2.

Table 2. The detailed characteristics of datasets.

| Dataset | #Instances | #Features | #Classes |
| --- | --- | --- | --- |
| Glass | 214 | 9 | 6 |
| Breast Cancer | 699 | 9 | 2 |
| Heart | 270 | 13 | 2 |
| Wine | 178 | 13 | 3 |
| German | 1000 | 24 | 2 |
| Ionosphere | 351 | 34 | 2 |
| Sonar | 208 | 60 | 2 |
| Hill-valley | 1212 | 100 | 2 |
| Musk1 | 476 | 167 | 2 |
| Arrhythmia | 452 | 279 | 16 |
| LSTV | 126 | 310 | 2 |
| Isolet5 | 1559 | 617 | 26 |

As shown in Table 2, two small, five medium, and five large datasets are used for the experiments.

To evaluate the performance of the proposed approach against the other methods, we perform qualitative evaluation by Pareto front investigation and quantitative evaluation by C-metric and hypervolume (HV) indicator. The C-metric calculates the dominant relationship between two Pareto front solutions by Equation (3) [3].

$$C(P_1, P_2) = \frac{|\{p_2 \in P_2 | \exists p_1 \in P_1 : p_1 \geq p_2\}|}{|P_2|} \quad (3)$$

In Equation (3), $C(P_1,P_2)$ represents the portion of solutions in Pareto front $P_2$ that are dominated by at least one solution in Pareto front $P_1$. The value of $C(P_1,P_2)$ is between zero and one. If $C(P_1,P_2)=1$, at least one solution in $P_1$ dominates all solutions in $P_2$. However, $C(P_1,P_2)=0$ indicates that no solution in $P_2$ is dominated by a solution in $P_1$. A higher value of $C(P_1,P_2)$ indicates that the method with the Pareto front $P_1$ outperforms the other. The C-metric measures only the convergence of a method to the final Pareto front. However, the HV indicator is a popular metric that measures both diversity and convergence to the final Pareto front [3]. It calculates the dominated region created by the Pareto front bounded by a reference point in the objective space. A higher value of HV indicates better diversity and convergence of the Pareto front.

Figure 1 shows the Pareto front analysis of different methods on different datasets. For each method, we performed ten independent runs and reported the best results.

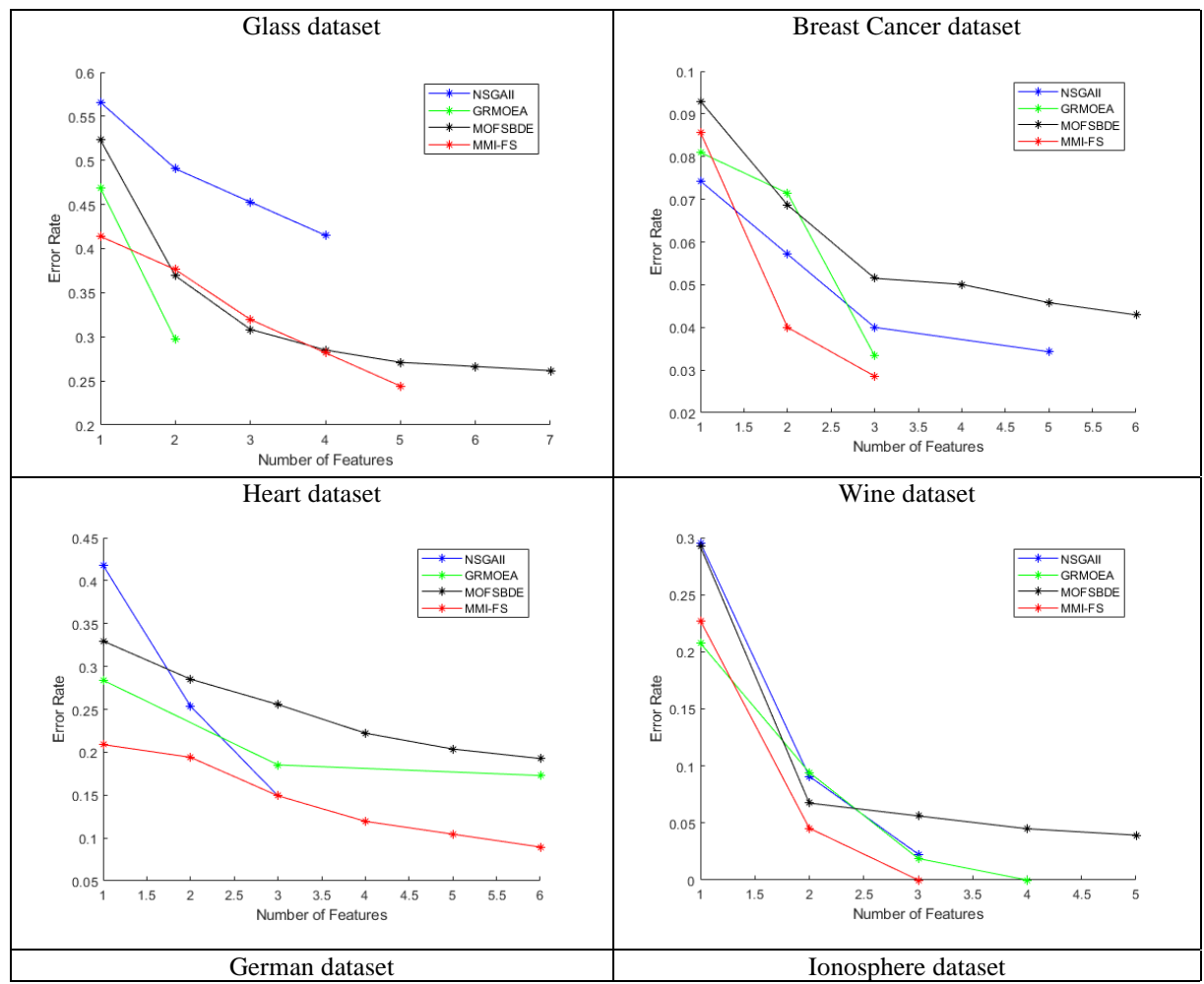

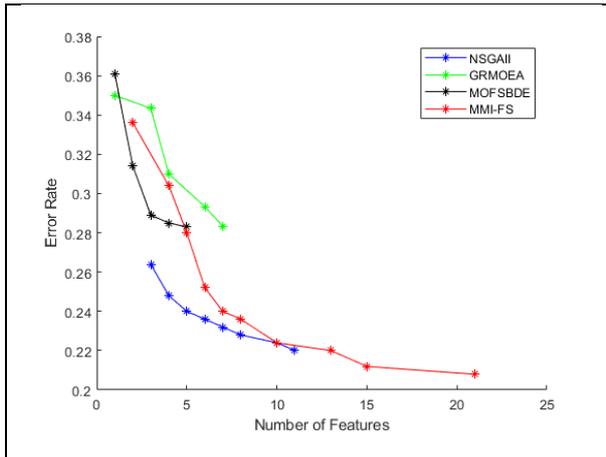
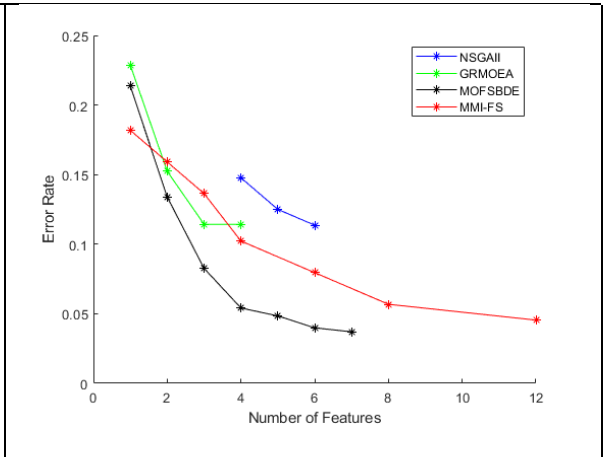

Sonar dataset | Hill-valley dataset

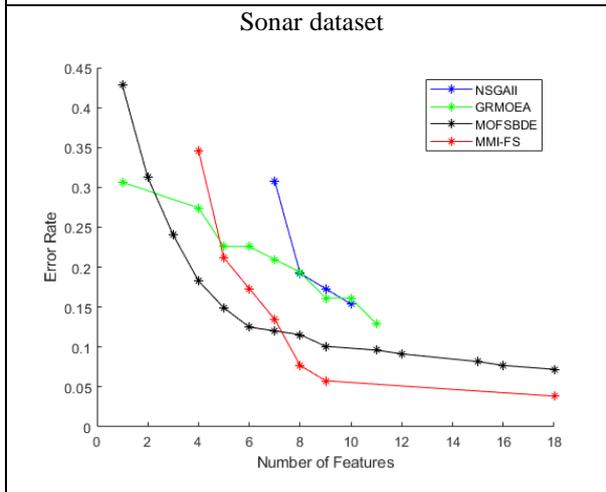
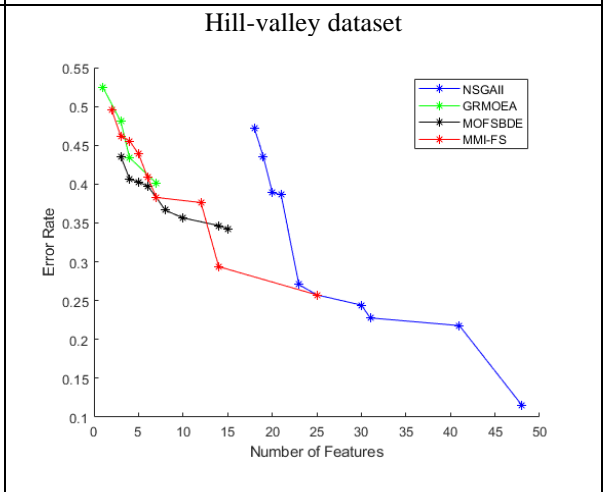

Musk1 dataset | Arrhythmia dataset

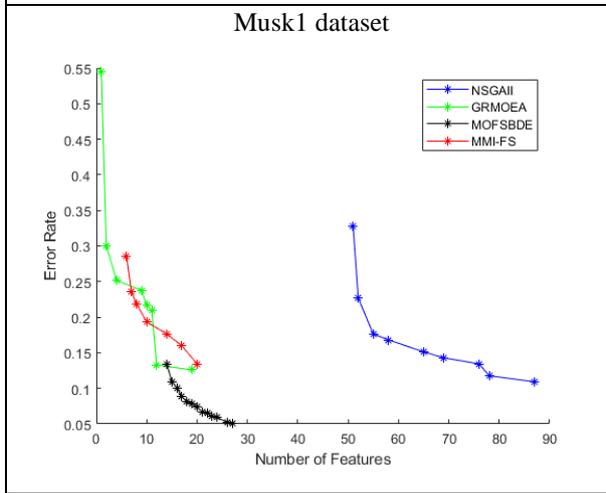
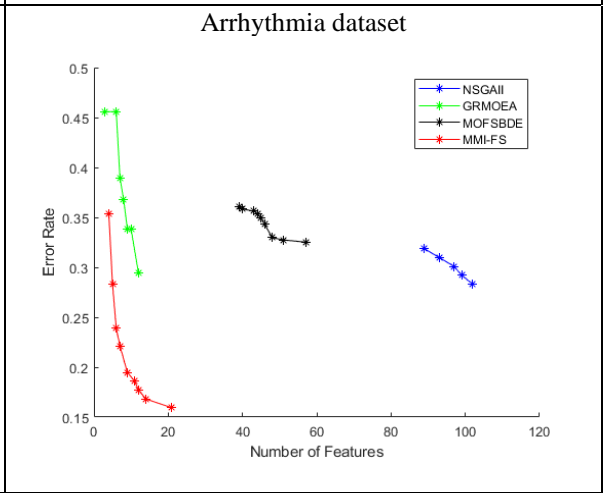

LSTV dataset | Isolet5 dataset

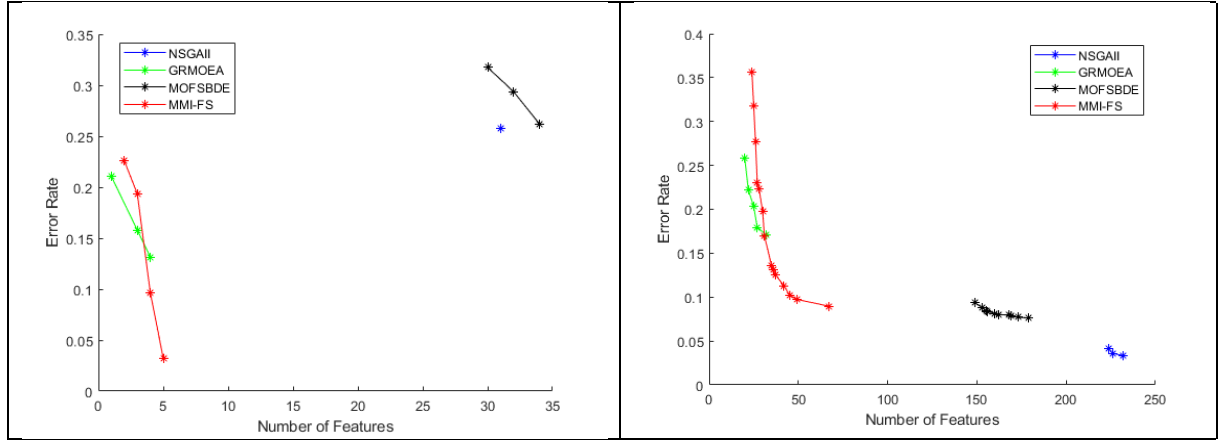

**Fig.1. The Pareto front achieved by different methods.**

As shown in Figure 1, almost all solutions of the proposed approach dominate the other methods in Arrhythmia and Heart datasets. For each of the Breast Cancer and Wine datasets, there is only one point that have greater values than those in other methods. For the remaining dataset, MMI-FS reported improvements in comparison to other methods.

Table 3 compares the C-metric values *C(MMI-FS,X)* and *C(X,MMI-FS)* for different methods on different datasets.

**Table 3: C-metric values between MMI-FS and other methods.**

|  | C(MMI-FS, X), X is: | | | C(X,MMI-FS), X is: | | |
|---|---|---|---|---|---|---|
| Dataset | NSGA-II | MOFSBDE | GRMOEA | NSGA-II | MOFSBDE | GRMOEA |
| Glass | 1 | 0.7143 | 0.5000 | 0 | 0.4000 | 0.4000 |
| Breast Cancer | 0.7500 | 1 | 0.6667 | 0.3333 | 0 | 0.3333 |
| Heart | 0.4000 | 1 | 1 | 0 | 0 | 0 |
| Wine | 1 | 1 | 0.7500 | 0 | 0 | 0.3333 |
| German | 0 | 0.2000 | 0.8000 | 0.6000 | 0.2000 | 0 |
| Ionosphere | 1 | 0.1429 | 0.5000 | 0 | 0.8571 | 0.2857 |
| Sonar | 1 | 0.5000 | 0.7778 | 0 | 0.5714 | 0.1429 |
| Hill-valley | 0.5000 | 0.2222 | 0.5000 | 0 | 0.5556 | 0.2222 |
| Musk1 | 0.7778 | 0 | 0.3750 | 0 | 0.4286 | 0.5714 |
| Arrhythmia | 1 | 1 | 0.8571 | 0 | 0 | 0 |
| LSTV | 1 | 1 | 0.3333 | 0 | 0 | 0.5000 |
| Isolet5 | 0 | 0.1000 | 0.2000 | 0 | 0 | 0.4286 |

As shown in Table 3, for almost all datasets, the proposed MMI-FS reports high values of *C(MMI-FS, X)* and low values of *C(X,MMI-FS),* which means the superiority of the proposed algorithm. However, there are some cases that the solutions of the two methods are not superior to each other. For example, neither solutions in NSGA-II nor the proposed MMI-FS could dominate the solutions of each other for the Isolet5 dataset.

Although C-metric measures the convergence of the final Pareto front, it is needed to measure both diversity and convergence of the obtained Pareto front. For this purpose, we calculate the HV of different methods. Table 4 summarizes the HV values for different methods on different datasets. It should be noted that the reference point in our experiments is set to (100, *m*), where *m* is the number of features for a given dataset. Furthermore, for better comparison, Figure 2 represents the HV values shown in Table 2.

**Table 4. The Hypervolume values for different methods on different datasets.**

| Dataset | NSGA-II | MOFSBDE | GRMOEA | MMI-FS |
|---|---|---|---|---|
| Glass | 0.4904 | 0.6055 | 0.6060 | **0.6260** |
| Breast Cancer | 0.8501 | 0.8401 | 0.8496 | **0.8560** |
| Heart | 0.7864 | 0.7193 | 0.7433 | **0.8154** |
| Wine | 0.8758 | 0.8635 | 0.8985 | **0.9023** |
| German | 0.6767 | 0.6821 | 0.6762 | **0.7009** |
| Ionosphere | 0.7807 | **0.9243** | 0.8553 | 0.9097 |
| Sonar | 0.7438 | **0.8912** | 0.8381 | 0.8820 |
| Hill-valley | 0.6802 | 0.6337 | 0.5887 | **0.7053** |
| Musk1 | 0.6084 | **0.8677** | 0.8585 | 0.8294 |
| Arrhythmia | 0.4871 | 0.5795 | 0.6951 | **0.8261** |
| LSTV | 0.6670 | 0.6661 | 0.8650 | **0.9601** |
| Isolet5 | 0.6157 | 0.7003 | 0.8015 | **0.8715** |

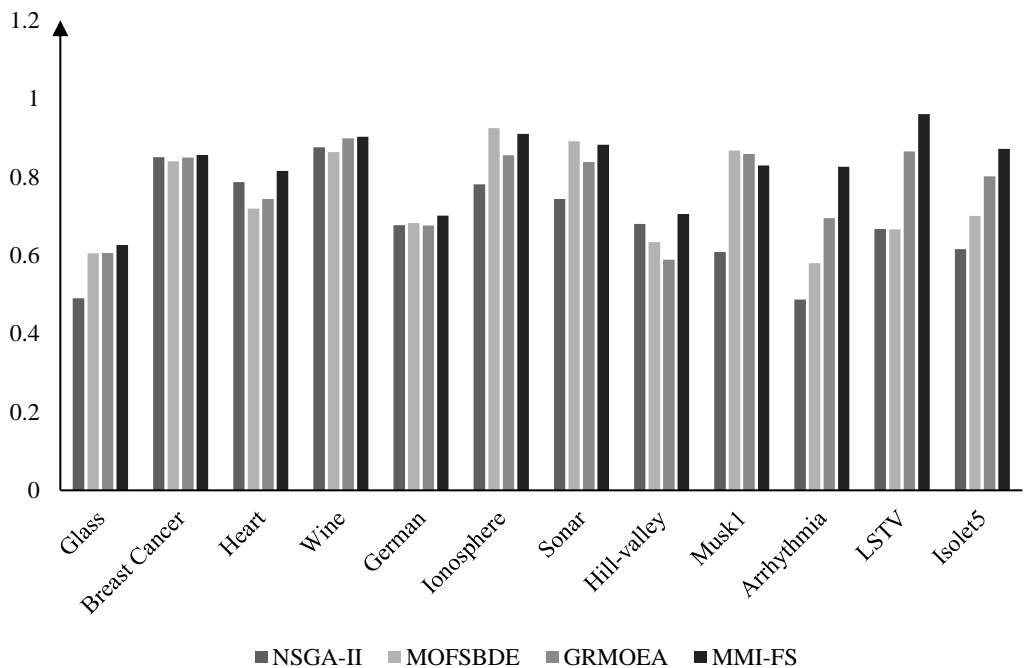

**Fig. 2. The Hypervolume values for different methods on different datasets.**

As it can be seen from Table 4 and Figure 2, MMI-FS outperforms the other methods on nine out of 12 datasets. For Ionosphere and Sonar datasets, the proposed method reports a smaller value of HV than the MOFSBDE method. Finally, for the Musk1 dataset, MMI-FS still shows an improvement in comparison with NSGA-II.

For efficiency analysis of the obtained results, we perform Wilcoxon signed-rank test [28] and Friedman test [7]. The significance levels for both tests are set to 0.05. It should be noted that the first one is used for pairwise performance comparisons between the proposed algorithm and the other methods. Table 5 represents the test results for HV metric.

**Table 5.** *p*-value of Wilcoxon signed-rank test in term of HV metric.

| Method | *p*-value |
|---|---|
| NSGAII | 0.00048828125 |
| MOFSBDE | 0.01611328125 |
| GRMOEA | 0.00488281250 |

As we conclude from Table 5, the proposed MMI-FS reports a significant difference.

We also used the Friedman test to assess the performance of all approaches on all datasets. The *p*-value for this test was equal to 0.0023, which proves that the overall performance of our proposed algorithm is significantly better than others in term of HV metric.

## 5. Conclusion and discussion

Feature selection is one of the most challenging phases for data preparation in the field of pattern recognition. However, it has several challenges that can be summarized as follows: firstly, increasing the number of features increases the size of the search space exponentially, and it makes this problem NP-hard. Consequently, the exhaustive search solution is unfeasible. To face this challenge, the metaheuristic techniques were widely applied to the FS problem. However, they also suffer from high time complexity. Secondly, the interaction between featured considerably affects the classification performance. A given feature may be relevant (irrelevant) to the target class by itself, but it may decrease (increase) the classification performance with some other features. To deal with the mentioned problems, we proposed a multi-objective feature selection approach based on the Pareto Archived Evolution Strategy (PAES) method. It is based on (1+1)-evolution strategy that reduces its computational cost and has several advantages such as simplicity and also its speed in exploring the solution space. Then, we integrated a novel conditional probability scheme into the proposed method. Using two introduced data structures, the proposed multi-objective FS method can handle the interaction between features. Furthermore, one of the main drawbacks of PAES is its blind mutation as the evolutionary operator, and, consequently, the generated offspring may have lower fitness than the parent. The introduced *winner* and *loser* individuals and also our probability scheme makes the proposed method capable to generated more promising offsprings. The obtained results on different real-world datasets and their statistical analysis showed a significant improvement compared to state-of-the-art methods. As a future direction for further study, we may work on a more promising mutation. Moreover, to have a better initialization for the proposed algorithm, it is a good idea if we firstly rank the features and select the top-ranked ones as the current and candidate individuals. Finally, we are also interested in applying the proposed multi-objective feature selection to different real-world applications.

## Compliance with ethical standards

**Conflict of interest** The authors declared no conflicts of interest with respect to the authorship and/or publication of this article.